\documentclass[a4paper, 10 pt, conference]{ieeeconf} 
\usepackage[dvipsnames]{xcolor,colortbl}
\usepackage{cite}
\usepackage{amsmath,amssymb,amsfonts}
\usepackage{algorithmic}
\usepackage{graphicx}
\usepackage{textcomp}
\usepackage{adjustbox,booktabs,multirow}
\usepackage{colortbl}

\usepackage{siunitx}
\usepackage{subfig}
\def\BibTeX{{\rm B\kern-.05em{\sc i\kern-.025em b}\kern-.08em
    T\kern-.1667em\lower.7ex\hbox{E}\kern-.125emX}}
\begin{document}

\title{\Large \bf 
Biasing Frontier--Based Exploration with Saliency Areas
}

\author{Matteo Luperto$^{1*}$, Valerii Stakanov$^{1,2}$, Giacomo Boracchi$^{2}$, Nicola Basilico$^{1}$,  and Francesco Amigoni$^{2}$% <-this % stops a space
\thanks{$^{1}$ Università degli Studi di Milano, Italy}
\thanks{$^{2}$ Politecnico di Milano, Italy}
\thanks{$^{*}$ Corresponding author: \tt\small matteo.luperto@unimi.it}}

\maketitle

\begin{abstract}
Autonomous exploration is a widely studied problem where a robot incrementally builds a map of a previously unknown environment. The robot selects the next locations to reach using an exploration strategy. To do so, the robot has to balance between competing objectives, like exploring the entirety of the environment, while being as fast as possible. Most exploration strategies try to maximise the explored area to speed up exploration; however, they do not consider that parts of the environment are more important than others, as they lead to the discovery of large unknown areas. We propose a method that identifies \emph{saliency areas} as those areas that are of high interest for exploration, by using saliency maps obtained from a neural network that, given the current map, implements a termination criterion to estimate whether the environment can be considered fully-explored or not. We use saliency areas to bias some widely used exploration strategies, showing, with an extensive experimental campaign, that this knowledge can significantly influence the behavior of the robot during exploration.
\end{abstract}

%\begin{IEEEkeywords}
%component, formatting, style, styling, insert
%\end{IEEEkeywords}

\section{Introduction}
%\red{check if progress of gradcam is semantically relevant during exploration}
%\red{there was a problem with references with images, two images had the same label. it should be fixed now but we need to doublecheck all refs}
Exploration, a widely studied problem within the field of autonomous mobile robotics, involves a robot building a \emph{map} of an unknown environment by iteratively planning and executing a sequence of actions. While some learning-based approaches have been recently proposed~\cite{cao2023}, exploration is customarily conducted by following a Next Best View (NBV) approach\cite{journalsijrrGonzalez-BanosL02}: given the current map, the robot computes a set of candidate locations to reach and selects, according to an \emph{exploration strategy}, the most promising one. In this work, we consider \emph{frontier}s, namely the boundaries between the known and the unknown space, as candidate locations to reach in order to make progress in exploration \cite{yamauchi1997frontier}. After the robot reaches the most promising frontier, it integrates the acquired knowledge into its map and repeats the process until a \emph{termination criterion} is met.

%\red{Add citations here, and we can reduce this part}
\begin{figure}[t]
\centering
\includegraphics[angle=90,width=0.55\linewidth]{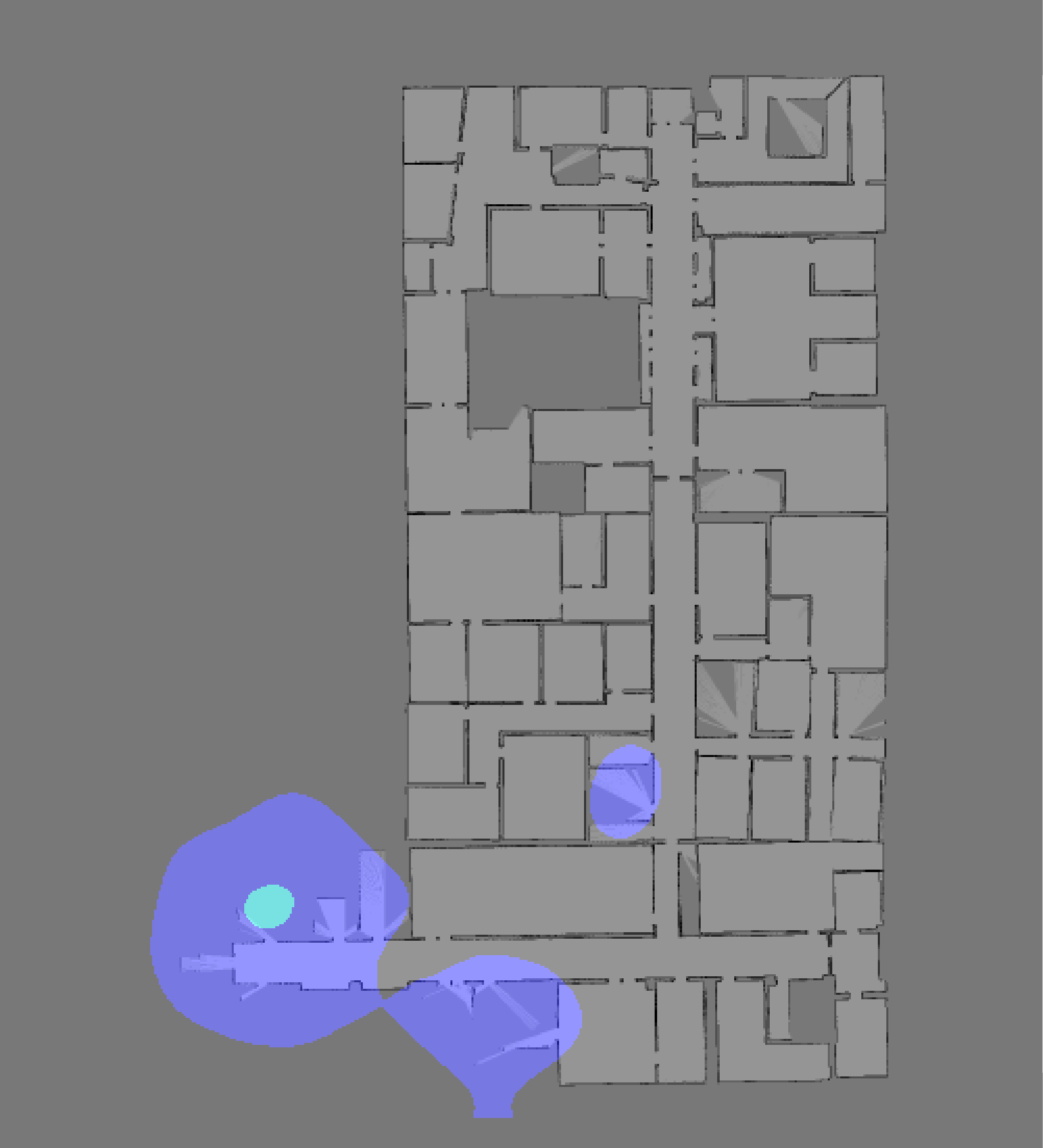}
\caption{Not all frontiers are equally important for exploration; some can lead to locally finishing the exploration of a room, while others can lead the robot to an entirely different part of the building. We identify the \emph{saliency areas} as areas of high interest for exploration, highlighted in light blue, and we integrate this idea into different exploration strategies.}
\label{fig:example}
\end{figure}

The goal of exploration is to eventually build a map of the environment~\cite{lluvia2021active}.
%that is as accurate as possible
 However, to obtain such a goal, the robot needs to balance between at least two competing objectives. On the one hand, the map should cover the entirety of the environment. On the other hand, there is a need to explore the environment efficiently, thus minimising the exploration time and the total path length of the robot.
Within this context, several works \cite{ericson2021understanding,ericson24,placed2022enough,RAS2020,ericson2025information} have shown how balancing these two needs is far from straightforward.
Exploration strategies that prefer to explore first the surroundings of the robot's location, such as \cite{yamauchi1997frontier}, increase systematically but slowly the mapped area. %However, their behaviour is locally greedy and far from an optimal exploration strategy, as they do not maximise the explored area with time.
A more aggressive exploration of the environment selects highly-promising locations that may be far away from the robot current location and can result into a high percentage of mapped area in early stages of exploration, but also in leaving behind several partially explored regions of the environment, thus forcing the robot to \emph{backtrack} to previously visited locations in the late stages of the process \cite{ericson2021understanding}.
%Eventually, strategies that prefer early-stage fast exploration require longer to complete the exploration time when compared to strategies that perform greedy exploration\cite{journalsijrrGonzalez-BanosL02}.

%Exploration strategies that prefer to always reach the nearest frontier first \cite{yamauchi1997frontier}, thus not considering the information gain, can build a map that covers most of the environment, but are usually slow as they exhaustively explore the surroundings of the robot location before moving to more interesting and informative frontiers. Exploration strategies that prefer the selection of frontiers with high information-gain (while also considering their distance) are often very fast in exploring most of the areas of the building, but then force the robot to \emph{backtrack} to previously visited locations where some low-information-gain frontiers have been left behind. This behaviour is more evident when the accuracy of the estimation gain increases, as an example when machine learning methods are used to estimate the information gain \cite{shrestha2019learned}.

% rephrase
The choice of which exploration strategy is better to use is also implicitly influenced by the termination criterion \cite{luperto2024estimating}, which decides when an exploration process should be considered finished. In some applications, the exploration is stopped when the process is no longer acquiring relevant information, for example, when no frontiers exceeding a minimum length can be explored~\cite{placed2022enough}. This is usually done when the main goal is to have full coverage of the map. In such cases, it is preferable to use exploration strategies that explore first the robot's surroundings before going to other parts of the map. In other operational contexts, such as Urban Search and Rescue \cite{liu2013robotic}, the exploration process stops after a pre-defined (usually short) amount of time since the application requires quick delivery of (possibly incomplete) information about the environment to the human rescuers. In those settings, exploration strategies that try to reach first highly-informative frontiers, even if they are far away from the robot's current location, are commonly employed.
%Although the need for stopping criteria is well recognised, identifying robust termination criteria is still an under-investigated challenge\cite{cadena2016,placed2022enough,placed2023survey}. 

In most works, the exploration strategy used by the robot and the termination criterion employed to stop the exploration run are considered as two independent methods. As a result, the robot's behaviour is not affected by the fact that the robot is close (or not) to the end of the exploration.

Some exploration strategies estimate the quantity of new area that is visible by the robot when reaching a frontier, under the assumption that the frontiers that lead to the biggest expected increase in the mapped area are a better choice for the robot. This estimation is commonly called \emph{information gain}. Selecting high information gain frontiers drives the robot towards large frontiers in open regions, thereby increasing the likelihood of rapidly expanding the mapped area. %This mechanism has the benefit that it is both easy to compute, robust to different settings, and can empirically speed up the exploration process, especially in early stages of exploration runs.
However, while the information gain takes into account the expected amount of observable area from a target location, it does not take into account the fact that not all the areas of the environment are equally important \cite{li2015semantically}. As an example, two frontiers, one located in a corridor and the other in a small room, can have a similar  information gain; however, exploring the corridor first can lead the robot to a whole new unexplored part of the environment, thus allowing the discovery of other highly informative frontiers. At the same time, the exploration of the frontier inside the room is less likely to lead the robot to discover new frontiers. 

In \cite{luperto2024estimating}, we have presented a method that stops the exploration process if the estimated unmapped portions of the environment are deemed as uninformative. This is a common situation in late stages of exploration runs, where the robot requires a long time to explore small corners or portions of the environment that cover only $1$ or $2\%$ of its total area, without adding additional knowledge to the mapping process. %(as in Fig. \ref{fig:exp/progress}). 

In this paper, we originally use a by-product of the method \cite{luperto2024estimating} to estimate which portions of the map are most relevant to classify the map as not explored.
Intuitively, when there are still multiple frontiers left to explore, we estimate the progress of the exploration process and highlight those portions of the maps that are still relevant to explore.  We call these areas \emph{saliency areas}.  An example is shown in Fig. \ref{fig:example}: while there are still several frontiers across the whole environment, we identify that the corridor at the bottom of the map (in light blue) is the most promising one to explore to reach new large unexplored parts of the building. 
%While there are multiple frontiers inside different rooms, our method can detect that the exploration run is still far from over and highlights (in blue) the most promising location to continue the exploration, as in the corridor at the bottom of the map. We label this knowledge as \emph{saliency areas} (Section \ref{sec:estimation}).

%Saliency areas can enable the robot to identify which are the parts of the map that are more relevant to explore to reach the goal of ending the exploration run, thus triggering the termination criterion.
%The goal of this work is thus to empirically evaluate whether the knowledge of the most-important areas of the map that need to be explored can inform different exploration strategies. 

We thus propose a method to bias an exploration strategy using the knowledge retrieved from the termination criterion, thus allowing the two methods to influence each other. In this way, we modify exploration strategies by adapting their choices to the status of the exploration process. We present counter-intuitive experimental results that, with certain exploration strategies, adding a negative bias (a cost to pay) to visit highly promising frontiers can be a better choice for the robot.
%To do so, we identify with the method from \cite{luperto2024estimating} which are the most promising portions of the map left to explore, and we introduce an additional component to evaluate each frontier that estimates how much each frontier is relevant to bring the whole exploration process to its conclusion. 

Our contributions are thus the following: (1) we present a method to extract saliency areas of interest for exploration, by using a by-product of the method~\cite{luperto2024estimating}; (2) we extend different exploration strategies to use the concept of saliency areas; (3) we provide an extensive experimental evaluation using simulated environments of how the knowledge of saliency areas can influence and change the behaviour of widely used exploration strategies.

\section{Related Work}\label{sec:related work}

%\red{\cite{ho2024mapex,akai2023slamer,harutyunyan2025mapexrl,baek2025pipe,p2explore}+ other papers referenced in latest ericson's paper.}

%While the general goal of an autonomous mobile robot exploring initially unknown (static) environments is to obtain their complete maps, there are often additional desiderata to address, such as maximizing the explored area within a time budget \cite{amigoni2013evaluating} or maximizing the quality of the reconstructed map \cite{placed2023survey}. In this context, the main (and most studied) issue a robot faces is to decide which is the next location to reach to progress the exploration  \cite{lluvia2021active,stachniss2005mobile}.
%Over the years, different exploration strategies have been proposed \cite{stachniss2009robotic}. A common one is to identify a few candidate locations at the boundaries between the explored and unexplored portions of the environment, and to select the most promising location to reach, according to some exploration strategy. These boundaries are called frontiers, and their ranking is commonly based on their expected \emph{information gain}, i.e., an estimate of the amount of knowledge to be acquired by the robot when visiting the locations, and on their \textit{distance} from the current location of the robot \cite{yamauchi1997frontier,lluvia2021active, amigoni2017multirobot}. 
The main goal of autonomous exploration is to acquire a map of an initially unknown environment, usually considered static. A recent survey \cite{brugali2025mobile} covers well the broad literature within this field. Along with the
%primary 
goal to have a complete map, there are often additional desiderata, such as maximising the quality of the map built by the robot \cite{placed2023survey} or the amount of mapped area within a time budget \cite{amigoni2013evaluating}. 

%To acquire these goals, the robot actively performs a set of actions, each consisting of moving to a new location to make a new observation \cite{lluvia2021active}. %Each action is usually planned independently from previous and future ones, and using the knowledge available at the given time. 
%The strategy used by the robot to make such a choice is called an exploration strategy.

%Frontier-based exploration is a family of common exploration strategies, where the robot identifies a few candidate locations at the boundaries between the explored and unexplored portions of the environment, and selects the most promising location to reach, according to some exploration strategy. These boundaries are called frontiers. %and their ranking is commonly based on their expected \emph{information gain}, i.e., an estimate of the amount of knowledge to be acquired by the robot when visiting the locations, and on their \textit{distance} from the current location of the robot \cite{yamauchi1997frontier,lluvia2021active, amigoni2017multirobot}

%The goal of exploration is to eventually build a map of the environment that is as accurate as possible \cite{lluvia2021active}. However, to obtain such a goal, the robot needs to balance between two competing objectives; one the one hand,  the map should cover the entirety of the area of the environment. On the other hand, there is a need to explore the environment as fast as possible, thus minimising the exploration time and the total path length of the robot. 
Exploration strategies commonly rank the frontiers by using a combination of distance from the robot's current position (a cost the robot needs to pay) and its expected information gain (IG), namely an estimate of the novel information about the environment that can be acquired by the robot when visiting a frontier (a gain that the robot wants to maximise).
%However, their behaviour is locally greedy and far from an optimal exploration strategy, as they do not maximise the explored area with time.
%Eventually, strategies that prefer early-stage fast exploration require longer to complete the exploration time when compared to strategies that perform greedy exploration\cite{journalsijrrGonzalez-BanosL02}.
However, several works \cite{ericson2021understanding,ericson24,placed2022enough,RAS2020,ericson2025information} have shown how balancing these two needs is far from straightforward. Exploration strategies that prefer to always reach the nearest frontiers \cite{yamauchi1997frontier}, thus not considering the information gain, can build a map that covers most of the environment, but are usually slow as they exhaustively explore the surroundings of the robot location before moving to more interesting and informative frontiers. Exploration strategies that prefer the selection of frontiers with high information gain (while also considering their distance) are often very fast in exploring most of the areas of the building, but then force the robot to backtrack to previously visited locations where frontiers with low information gain have been left behind.
Eventually, strategies that prefer early-stage fast exploration can require a longer time to fully explore the environment than strategies that perform a more systematic exploration\cite{journalsijrrGonzalez-BanosL02}. %This behaviour is more evident when the accuracy of the estimation gain increases, as an example when machine learning methods are used to estimate the information gain \cite{shrestha2019learned}.

Early works estimated the information gain of each frontier using optimistic approaches, such as assuming that all the unmapped area beyond the frontier is free~\cite{bircher2016}, or by using some heuristics, such as considering longer frontiers as more promising, i.e., with higher information gain \cite{journalsijrrGonzalez-BanosL02,amigoni2008icra}.

Several recent methods \cite{luperto2021exploration,katsumata2022map,shrestha2019learned,ericson24,SEER,ho2024mapex,baek2025pipe,p2explore} focus on \emph{predicting} a plausible (yet possibly inaccurate) status of the occupancy of the unobserved parts of the environment given its partial 2D map. While the details of the methods differ significantly, their general approach can be summarized as follows: the current map obtained by the robot is fed to a model, which exploits the fact that indoor environments present some regularities \cite{luperto2013}. The model is then used to predict a plausible, yet possibly inaccurate, state of the portions of the environment that are not mapped by the robot. The predicted map is used to inform the exploration process as it allows the robot to better estimate the information gain of each frontier. As a result, the robot is faster in exploring a large portion of the area during the early steps of exploration, but at the expense of more backtracking at later stages.
The prediction of the map is usually done using deep learning inpainting methods, such as in \cite{katsumata2022map,shrestha2019learned,ericson24,SEER,ho2024mapex,baek2025pipe,p2explore}, or feature-based ones \cite{luperto2021exploration}. To cope with the fact that the predicted map can be inaccurate, some methods produce different estimates of possible configurations of the same environment, and use them to get a probabilistic estimate of the information gain \cite{ho2024mapex}. A limitation of these works is that, as predicting the unseen parts of a map is difficult in real environments or with noisy sensor readings, most of these work are designed, trained, and tested in clean maps of empty environments assuming perfect sensing, navigation, and localization of the robot \cite{shrestha2019learned,ho2024mapex,baek2025pipe,p2explore}.

%We also use deep learning models and we leverage the visual representation of the current explored map, but we solve a significantly different problem: we directly use the currently available map to estimate the status of the exploration mission, without predicting the occupancy of the unobserved parts of the environment.

%Differently from the many studies on exploration strategies, some works highlight that deciding when to stop exploration is a particularly relevant, yet under-investigated, problem \cite{cadena2016,placed2022enough,placed2023survey}. 
%In \cite{placed2022enough}, an overview of the stopping criteria that are commonly used to end exploration is presented, as well as a discussion of their limitations. 
%A straightforward criterion is to stop the exploration process when a time budget is reached \cite{leung2008active}; however, the time budget should be tuned for each environment and exploration strategy, and there is no guarantee that the map obtained by the robot when the time budget expires is complete.

While most research has focused on exploration strategies, some works have pointed out that deciding when to stop exploration is a particularly relevant yet under-investigated problem \cite{placed2022enough,placed2023survey}. In \cite{placed2022enough}, the authors present an overview of common stopping criteria and discuss their limitations. A simple approach is to stop exploration once a pre-set time budget expires \cite{leung2008active}, but this method offers no guarantee that the resulting map is complete. Other works propose ending exploration when no candidate locations remain \cite{yamauchi1997frontier}, which ensures map completeness but often results in long exploration times. %as the robot spends considerable effort reaching faraway, less interesting areas \cite{ericson2021understanding,RAS2020}. %This inefficiency is typical of exploration strategies that predict unobserved parts of the environment to estimate information gain \cite{luperto2021exploration,katsumata2022map,shrestha2019learned,RAS2020}, where, as noted by \cite{ericson2021understanding}, a robot can quickly map most of the environment but then spend up to 71\% of its total time covering marginal details representing only about 10\% of the area.
%Some methods stop exploration once a certain percentage of the environment is mapped \cite{amigoni2013evaluating}, yet this requires prior knowledge of the environment’s size or the ability to infer it, and high area coverage does not necessarily imply that all important regions have been observed. 
%Other approaches compare maps generated at different times and stop when they are nearly identical \cite{amigoni2018improving}, signaling that no new areas are being discovered, although this technique may fail when the robot is in transit between distant frontiers. 
Some methods stop exploration once a certain percentage of the environment is mapped \cite{amigoni2013evaluating}, thus requiring the knowledge of the total size of the environment, or when maps generated at different times are nearly identical \cite{amigoni2018improving}, signaling that no new area has been observed.
In these works, the exploration strategy is not affected by the fact that the robot is close (or not) to reaching the end of exploration.

The termination criterion is intertwined with the exploration strategy in information-theoretic methods, such as those based on map entropy \cite{ghaffari2019sampling,placed2022enough}; those methods propose stopping exploration when the entropy falls below a threshold or saturates, and the same entropy measure is used to drive exploration.
%To address some of these limitations, \cite{placed2022enough} proposes using the Theory of Optimal Experimental Design (TOED), employing a D-optimality criterion to define a stopping condition based on the statistical utility of the map; exploration is halted once no significant new information is added. 
The tight link between the termination criterion and the exploration strategy makes it difficult to combine these approaches with other strategies like frontier-based exploration.

In our work, we do not use any prediction of the map to estimate the information gain, but we use the knowledge retrieved from a termination criterion to inform the exploration strategy. As our method relies only on the availability of the map, it can be combined with different exploration strategies.

\section{Proposed Method}\label{sec:method} \label{sec:proposed solution}

\begin{figure*}[t]
\centering
\subfloat[\label{fig:map30percent}]{\includegraphics[trim={0 0cm 0 0cm},clip,width=0.3\linewidth]{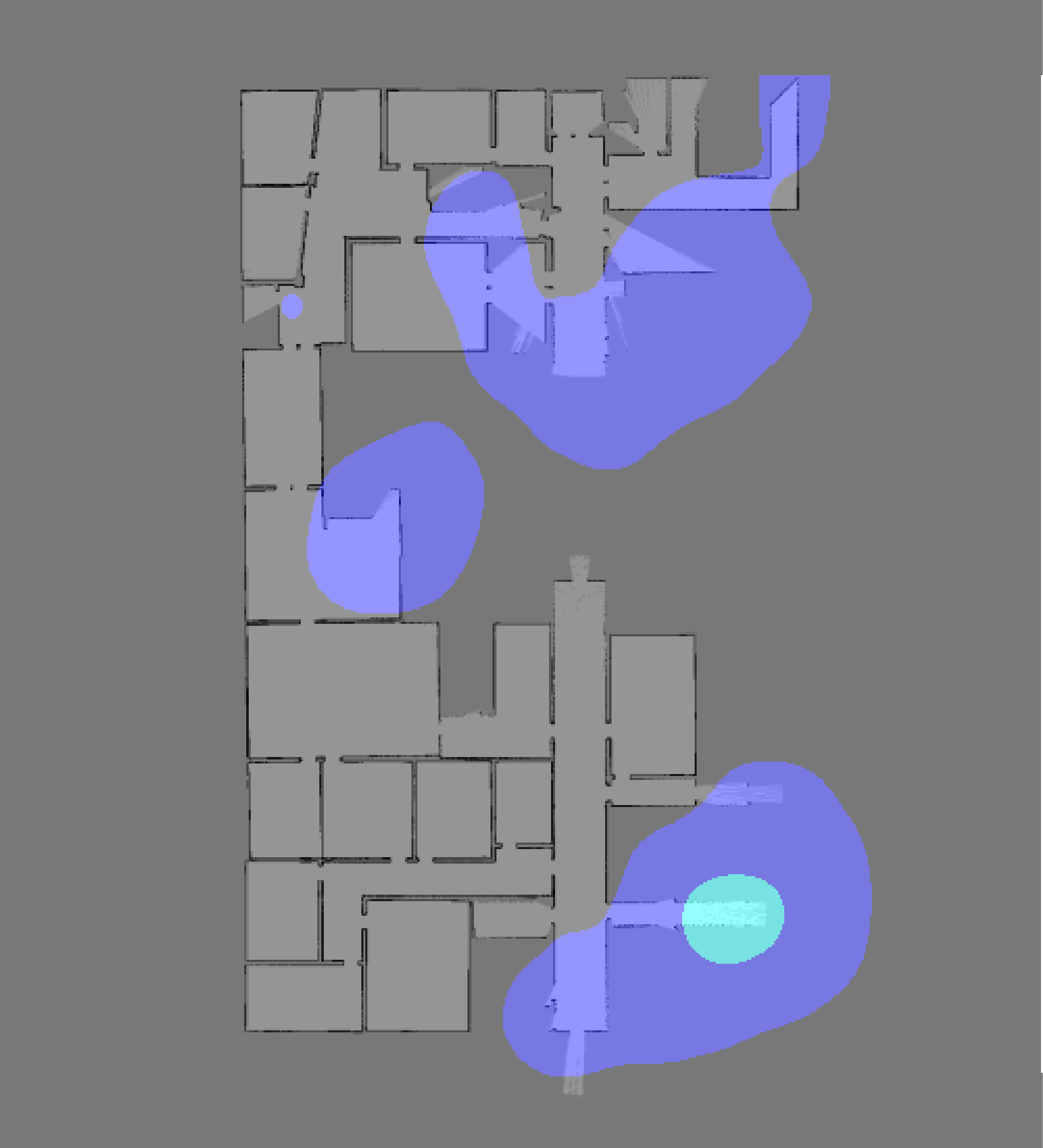}}
\subfloat[\label{fig:map100percent}]{\includegraphics[trim={0 0cm 0 0cm},clip,width=0.3\linewidth]{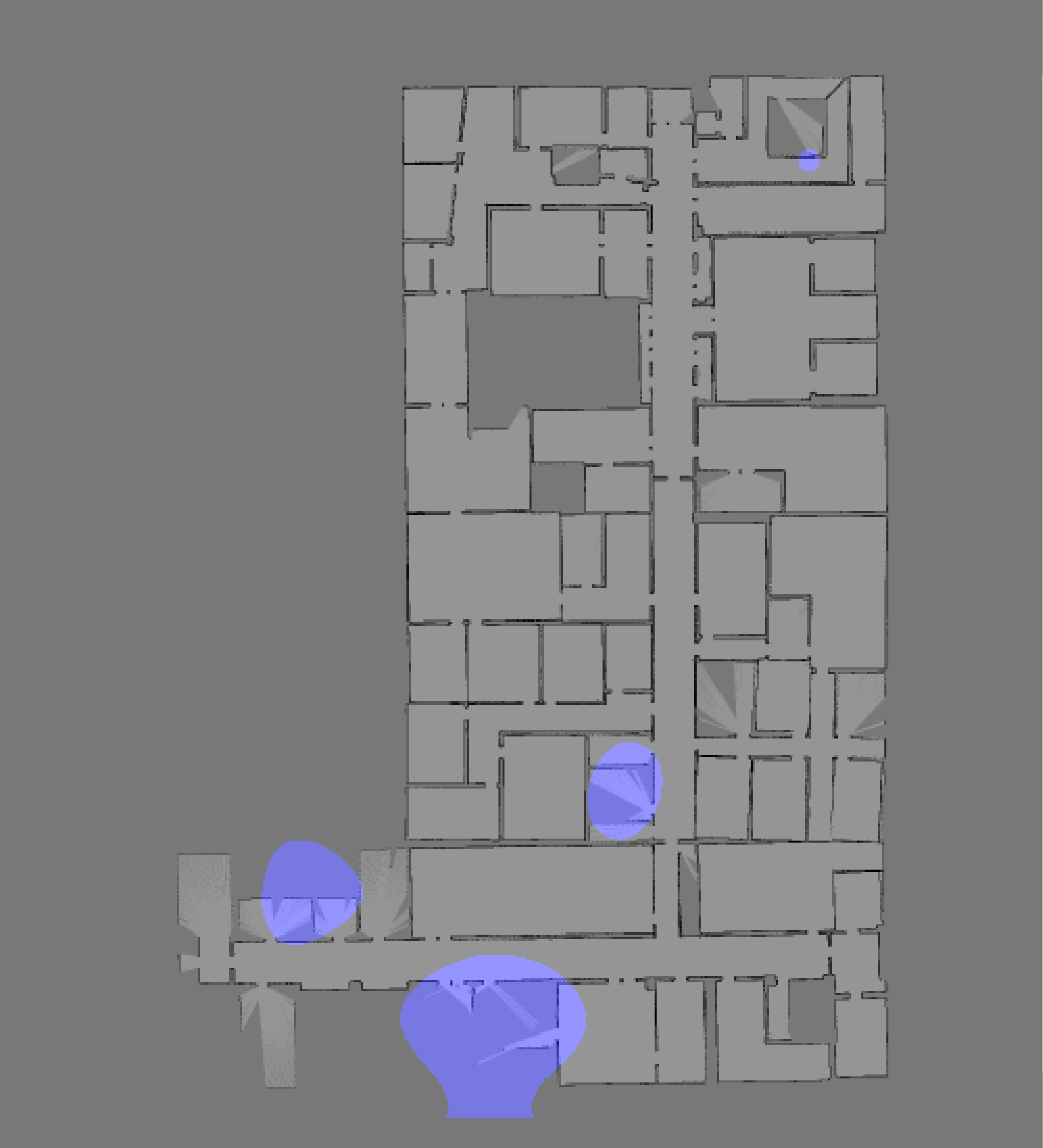}}
\subfloat[\label{fig:map1}]{\includegraphics[trim={0 0cm 0 0cm},clip,width=0.3\linewidth]{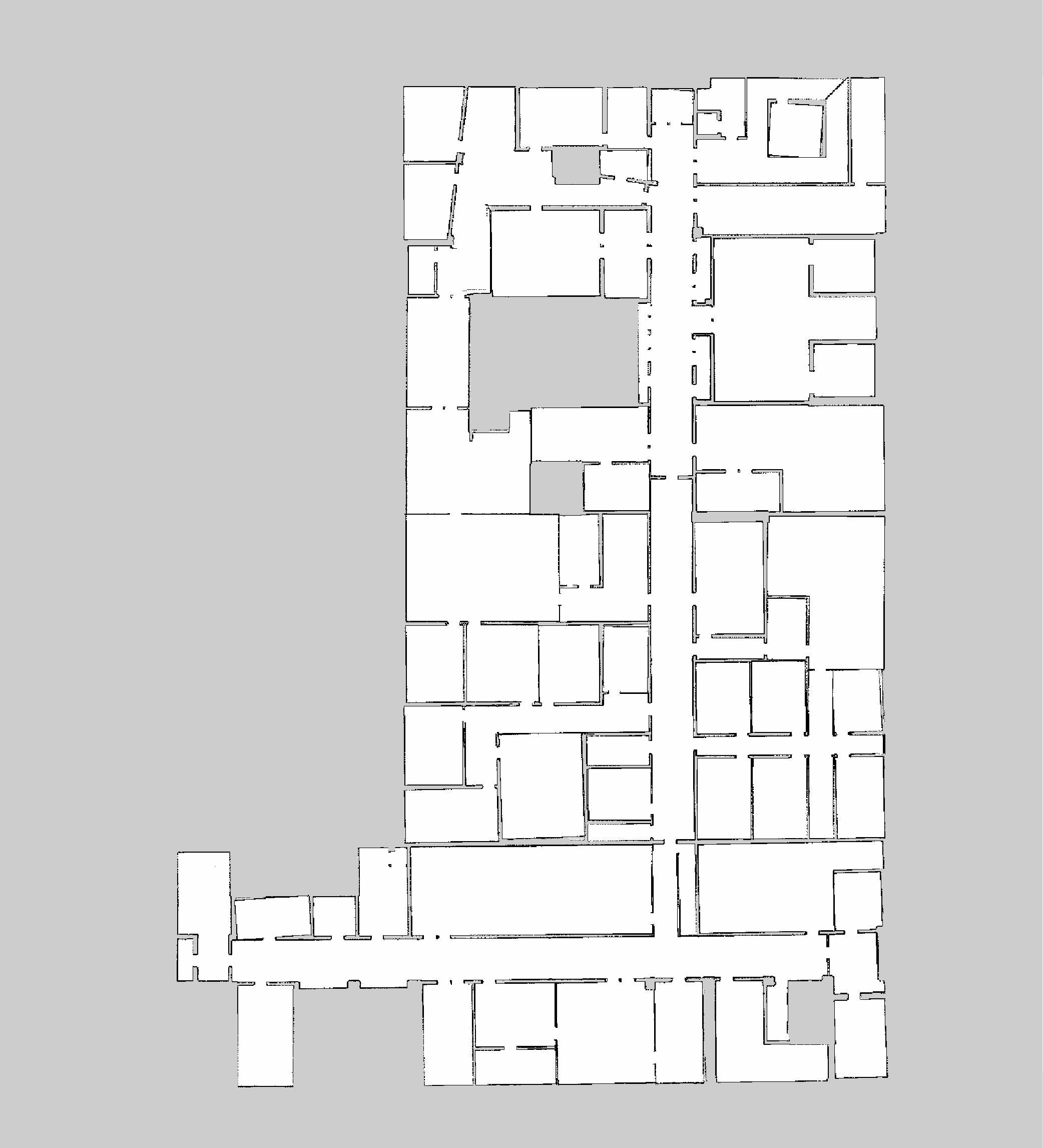}}
\caption{
Saliency areas at different stages of exploration for environment $e_1$. Full map in (c).}
\label{fig:exp/progress}
\end{figure*}

In this paper, we consider an autonomous mobile robot that explores an initially unknown two-dimensional environment using a frontier-based exploration strategy \cite{yamauchi1997frontier}. During exploration, the robot can access an evaluation of the current status of the exploration process, which is used as a termination criterion \cite{luperto2024estimating}.

The termination criterion identifies, within the map, \emph{saliency areas}. We detail this process in Section \ref{sec:estimation}.
Saliency areas denote the parts of the environment that are more relevant to explore to reach the goal of ending the exploration, thus triggering the termination criterion.
The goal of this paper is to investigate whether the knowledge of these areas can inform different exploration strategies. 

To do so, we focus on three different exploration strategies that represent different families of approaches to frontier-based exploration (Section \ref{sec:strategies}).  We then extend those strategies by adding an additional term to the frontier evaluation that represents its \emph{saliency} with respect to the exploration process as a whole (Section \ref{sec:integration}).

%\red{The IG alone it is not enough as it causes fast exploration but long backtracking, can we do better?}

\subsection{Frontier-based exploration}\label{sec:strategies}
A robot using a frontier-based exploration iteratively updates a 2D metric map $M$, which we assume to be a grid map, that represents the space that the robot can sense. Each cell of $M$ can be free, obstacle, or unknown.
The robot identifies the set $F$ of frontiers from the current map $M$. A frontier is a chain of free cells that are adjacent to unknown cells. The robot selects the best frontier $f^* \in F$ by maximizing a utility function:
\begin{equation} \label{eq:1}
u(f) = \alpha \cdot IG(f) - (1-\alpha) \cdot dist(f),
\end{equation}
where $IG(f)$ is the frontier's information gain, $dist(f)$ is its centroid's distance from the robot, and $\alpha \in [0,1]$ is a parameter that balances these two factors. Note that while $dist$ is a cost, that the robot wants to minimize, $IG$ is a gain to be maximized. Both $IG$ and $dist$ are normalized in $[0,1]$ using their maximum and minimum values for the frontiers in $F$.

According to this function, the exploration strategy prioritizes nearby frontiers that are also promising, as these are more likely to reveal large areas of unknown space. After the frontier $f^*$ is selected, the robot plans and follows a path to $f^*$'s centroid, updating the map with the new sensory data acquired while traveling.
We consider three different exploration strategies as representatives of different approaches. 

First, we consider a \emph{nearest-frontier} exploration strategy (\texttt{NF}) \cite{yamauchi1997frontier}, where $\alpha=0$ and thus the information gain is not considered by the robot when selecting the next frontier to visit. This strategy always selects the closest frontier. 

Then, we consider a second strategy, which uses the space that is potentially observable from the centroid of the frontier to estimate the information gain, as in \cite{journalsijrrGonzalez-BanosL02}, and thus we label it \texttt{IG}. Our estimate of the information gain is optimistic, as it assumes that the unobserved area beyond each frontier and within the robot's sensor footprint is free and can be all observed. This strategy, when compared to \texttt{NF}, quickly increases the area observed by the robot, but eventually results into a longer time required for the full coverage of the environment \cite{ericson2025information}. We set $\alpha=0.5$.

Finally, we consider a third strategy, named \texttt{IG$^*$}, that predicts the information gain of a frontier by estimating more accurately the status of the map beyond a frontier. This idea has been proposed by several recent approaches \cite{SEER,baek2025pipe,ho2024mapex,luperto2021exploration,ericson24} that predict the status of the unknown parts of the environment and use this prediction to estimate the area that is observable from the centroid of a frontier. In this way, the estimated information gain is more accurate and less optimistic than the one made by \texttt{IG}. Empirically, these methods allow a significant speed-up in the early stages of exploration runs, but at the cost of backtracking to observe frontiers with low information gain at the end \cite{ericson2021understanding}. We assume that the information strategy \texttt{IG$^*$} has the best possible knowledge, namely we use the full map of the environment to compute the exact information gain of each frontier, thus simulating a perfect map-prediction method. Also for \texttt{IG$^*$} we set $\alpha=0.5$.

\subsection{Estimating the progress of the exploration run}\label{sec:estimation}

The robot relies on the method presented in \cite{luperto2024estimating} and that uses an image depicting the current grid map to detect if it satisfactorily represents (or not) the whole environment. The classification output defines the termination criterion. We present here a brief overview of the method; please refer to \cite{luperto2024estimating} for full details.

The classification model is based on a convolutional neural network (CNN). The termination criterion is modeled as a visual recognition problem where the classification task is to distinguish between maps that are \texttt{not-explored} yet from those that are sufficiently \texttt{explored}. Starting from a  dataset of maps, we label as \texttt{not-explored} those maps that a human (who knows the full map of the environment) would consider incomplete, because they miss areas of interest, like entire rooms or relevant portions of the environment. We then consider as \texttt{explored} those maps that a human would consider complete enough to represent the full environment. Note that, in maps labelled as \texttt{explored}, some small parts of the environments (like corners) may not be entirely mapped yet, but such parts are not considered relevant and the mapping process can be stopped even if such portions are missing. %As noted in \cite{ericson2021understanding,placed2022enough,placed2023survey}, due to backtracking, exploring the final $<5\%$ of the area can take up to $70\%$ of the time, while possibly jeopardising the entire mapping process. 
Often, the map beyond these remaining frontiers can be easily estimated from the known part of the map\cite{luperto2021predicting,shrestha2019learned}, as in the case where a corner of the room has not been fully explored, but other walls and rooms close to it are already known. An example of an \texttt{explored} map is shown in Fig.~\ref{fig:rviz}.
%s noted in \cite{placed2022enough,placed2023survey}, the final stage of exploration are usually particularly expensive in terms of time required (see \cite{ericson2021understanding}) while adding a few square meters to the map and while possibly jeopardizing the entire mapping process. Then, often the missing details of the map can be estimated directly from the known part of the map, as in the case where a corner of the room has not been fully explored, but the other walls and rooms that are behind it are already part of the map. Finally, several recent works have shown how these small details of the map can be easily predicted using machine learning or feature based techniques \cite{luperto2021predicting,p2explore,shrestha2019learned,ho2024mapex,SEER}.
%An example of an (\texttt{not-})\texttt{explored} map can be found in Fig. (\ref{fig:map30percent}) \ref{fig:rviz}.

%The input for our method is a partial grid map $M_{t}$ built up to time $t$, where each cell can have three possible values: free (white), obstacle (black), or unknown (grey). The map is centred, scaled, and resized into a $500\times500$ pixels image. %Thus, the scaling factor is not fixed and depends on the size of the original map $M_{t}$. Therefore, our method can be applied to any grid map $M_{t}$ regardless of how it was obtained and our method can be used in combination with any exploration strategy.

We use as CNN an EfficientNet B1 \cite{tan2019efficientnet}, pre-trained on ImageNet \cite{deng2009imagenet} and fine-tuned on a dataset of about $20,000$ maps collected during different exploration runs \cite{amigoni2018improving}, which are automatically labelled as \texttt{explored} or not.

The output of the method is a label which indicates if the map is \texttt{not-explored} or \texttt{explored}, and thus if the exploration run should continue or can be stopped. 
We then leverage saliency maps on the trained CNN classifier to design new exploration strategies. Saliency maps, like Grad-CAM \cite{gradcam}, are explainability techniques that provide, for each input image, a heatmap that highlights which regions have primarily influenced the network to decide a specific class. Therefore, saliency maps associated with the class  \texttt{not-explored} can suggest which regions to prioritize in the exploration, as shown in Fig.~\ref{fig:example}. We call these regions \emph{saliency areas}.

%\textcolor{red}{IT SEEMS THAT THE REFERENCES TO THE FIGURES ARE MESSED UP. ALL REFERENCES MUST BE CAREFULLY DOUBLE CHECKED.}
These saliency areas evolve and change during exploration, as shown in Figs. \ref{fig:gradcampre},~\ref{fig:example}, and \ref{fig:exp/progress}a-b, that show four maps from the same exploration run. In early stages of exploration  (Fig. \ref{fig:gradcampre}), the model highlights most of the frontiers; with the progression of exploration (Fig. \ref{fig:exp/progress}a-b), only those frontiers that are in unexplored rooms or that led to new unobserved parts of the environment are highlighted. Finally, Fig. \ref{fig:exp/progress}c shows the map when it is labeled as \texttt{explored}. We can see how, in different stages of the exploration runs, saliency areas can highlight the most relevant portions of the environment to explore: while at early stages of exploration, all frontiers are equally relevant, in later stages only those that lead to new parts of the environment are highlighted.

\subsection{Biasing exploration with saliency maps}\label{sec:integration}

\begin{figure}[t]
\centering
\subfloat[\label{fig:gradcampre}]{\includegraphics[trim={0 0cm 0 0cm},clip,width=0.45\linewidth]{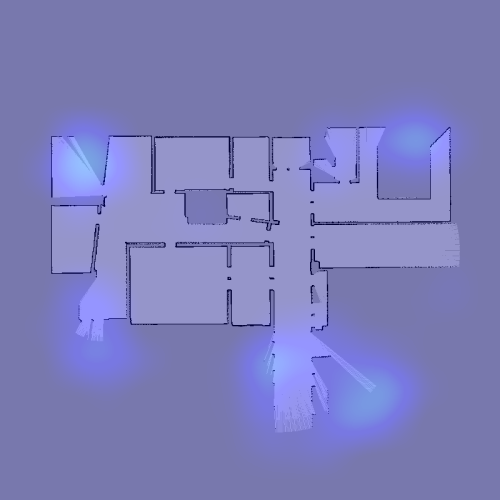}}
\subfloat[\label{fig:gradcampost}]{\includegraphics[trim={0 0cm 0 0cm},clip,width=0.45\linewidth]{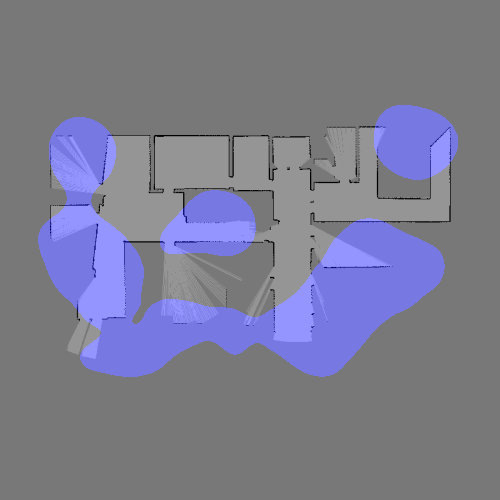}}
\caption{GRAD-Cam output (a) and saliency areas (b).}
\label{fig:exp/superimposing}
\end{figure}

In order to use saliency areas to inform different exploration strategies, we introduce a new term into the framework described in Eq. \ref{eq:1}, namely
\begin{equation} \label{eq:2}
u(f) = \alpha \cdot IG(f) - (1-\alpha) \cdot dist(f) +  \beta \cdot S(f),
\end{equation}
where $\beta$ is a weight and $S(f)$ evaluates the \emph{saliency} of the frontier.%, which is computed as described here and shown in Fig. \ref{fig:exp/superimposing}.

\begin{figure}[t]
\centering
\includegraphics[width=0.7\linewidth]{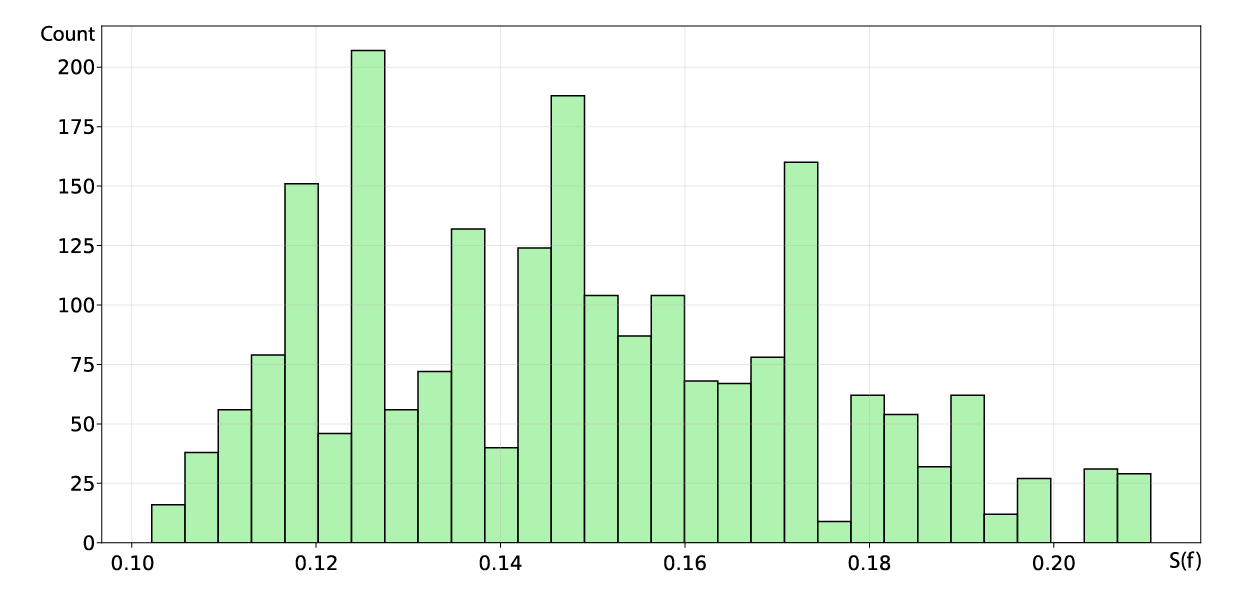}
\caption{Distribution of values of $S(f)$.}
\label{fig:intensity}
\end{figure}

The output of the Grad-CAM \cite{gradcam} is a mask of the same size as the image of the map with values in $[0,1]$, where the non-relevant areas have value $0$. An example of this mask is shown in Fig. \ref{fig:gradcampre}.
A threshold $\theta=0.1$ is applied to the Grad-CAM image, setting all values below  $\theta$ to $0$. Then, we identify all non-zero areas within the map using connected-component analysis, finding the shape of all saliency areas. The result of this thresholding step is shown in Fig. \ref{fig:gradcampost}. For each saliency area, we find the average value of its Grad-CAM and we use that for the entire area. Finally, $S(f)$ is computed by projecting the centroid of the frontier in the saliency map and using the value of the corresponding saliency area.
Fig. \ref{fig:intensity} shows the empirical distribution of $S(f)$ for a map, which is in the range $[0,0.25]$ and with an average of $0.15$.
% Overall, the average value of $S(f)$ for non-zero values is of $0.15$, and empirically the values are in a range $[0,2.5]$, as shown in Fig. \ref{fig:intensity}
Note that frontiers $f$ that are in a saliency area with $S(f)=0$ can still be visited by the robot and are not discarded. Moreover, the value $S(f)$ of the same frontier $f$ can change when the map is updated, and a new saliency map is computed.

\section{Experimental Evaluation}\label{sec:exp}
\begin{table*}[t!]
\adjustbox{max width=\textwidth}{%
\centering
\begin{tabular}{cl|ccccccc}
             Env.       & Strategy & $A_{30}$     & $A_{50}$     & $A_{70}$     & $A_{90}$    & $A_{95}$      & $A_{99}$      & $\bar{T}$       \\ \hline
\multirow{9}{*}{$e_1$} & \texttt{NF} &289,20 $\pm$ 118,98 & 535,60 $\pm$ 109,07 & 793,40 $\pm$ 131,18 & 1145,20 $\pm$ 193,76 & 1365,60 $\pm$ 331,53 & 1560,20 $\pm$ 256,52 & -  \\ 
& \texttt{NF+S(1)} &210,20 $\pm$ 43,81 & \textbf{393,60 $\pm$ 86,71} & \textbf{636,40 $\pm$ 124,14} & \textbf{935,00 $\pm$ 107,42} & 1245,80 $\pm$ 348,56 & 1603,40 $\pm$ 232,08 & 1389,20  \\ 
& \texttt{NF+S(2)} &244,20 $\pm$ 50,77 & 465,40 $\pm$ 100,21 & 724,80 $\pm$ 106,34 & 1210,60 $\pm$ 208,06 & 1367,80 $\pm$ 140,51 & 1657,40 $\pm$ 211,85 & 1478,60  \\
& \texttt{NF+S(4)} &210,60 $\pm$ 24,39 & 457,00 $\pm$ 83,54 & 754,80 $\pm$ 187,63 & 1241,20 $\pm$ 216,32 & 1479,80 $\pm$ 165,90 & 1731,60 $\pm$ 242,27 & 1560,60  \\
& \texttt{NF+S(-2)} &\textbf{198,20 $\pm$ 47,89} & 457,20 $\pm$ 169,46 & 646,00 $\pm$ 159,82 & 980,80 $\pm$ 286,77 & \textbf{1127,40 $\pm$ 241,33} & \cellcolor{CornflowerBlue} \textbf{1282,40} $\pm$ \textbf{226,39} & \textbf{1106,80}  \\ \cline{2-9}
& \texttt{IG} &198,20 $\pm$ 45,53 & 712,80 $\pm$ 533,56 & 1139,60 $\pm$ 514,49 & 1549,60 $\pm$ 437,18 & 1779,60 $\pm$ 586,78 & 2257,20 $\pm$ 801,66 & -  \\
& \texttt{IG+S(1)} &244,40 $\pm$ 46,01 & 403,20 $\pm$ 53,15 & \cellcolor{CornflowerBlue}\textbf{554,20 $\pm$ 52,29} & \cellcolor{CornflowerBlue}\textbf{855,80 $\pm$ 101,29} & \cellcolor{CornflowerBlue}\textbf{994,00 $\pm$ 168,52} & 1379,40 $\pm$ 184,71 & \cellcolor{CornflowerBlue}\textbf{1053,00}  \\ 
& \texttt{IG+S(2)} &257,20 $\pm$ 87,43 & 486,60 $\pm$ 116,52 & 674,80 $\pm$ 78,24 & 943,20 $\pm$ 104,29 & 1077,20 $\pm$ 105,42 & \textbf{1378,80 $\pm$ 267,88} & 1118,60  \\
& \texttt{IG+S(4)} &\cellcolor{CornflowerBlue} \textbf{189,00 $\pm$ 70,95} & \cellcolor{CornflowerBlue}\textbf{316,00 $\pm$ 60,66} & 575,40 $\pm$ 38,23 & 907,60 $\pm$ 139,18 & 1158,60 $\pm$ 269,09 & 1774,12 $\pm$ 327,47 & 1311,80  \\ \cline{2-9}
& \texttt{IG$^*$} &\textbf{204,00 $\pm$ 47,27} & \textbf{443,33 $\pm$ 108,51} & \textbf{704,67 $\pm$ 111,05} & 1297,33 $\pm$ 408,39 & \textbf{1410,33 $\pm$ 490,84} & 1847,83 $\pm$ 885,57 & -  \\
& \texttt{IG$^*$+S(2)} &230,00 $\pm$ 54,69 & 530,20 $\pm$ 163,05 & 830,80 $\pm$ 179,28 & \textbf{1233,40 $\pm$ 231,70} & 1488,40 $\pm$ 381,44 & \textbf{1742,80 $\pm$ 387,60} & \textbf{1586,20}  \\
& \texttt{IG$^*$+S(4)} &213,60 $\pm$ 45,07 & 581,00 $\pm$ 184,02 & 952,60 $\pm$ 253,16 & 1488,20 $\pm$ 461,78 & 1653,40 $\pm$ 409,25 & 1916,20 $\pm$ 572,55 & 1679,00  \\ \hline
\multirow{9}{*}{$e_2$} & \texttt{NF} &117,50 $\pm$ 59,71 & 160,33 $\pm$ 77,49 & 214,50 $\pm$ 59,46 & 407,50 $\pm$ 48,29 & 491,00 $\pm$ 65,01 & 580,83 $\pm$ 69,29 & -  \\
& \texttt{NF+S(1)} &56,20 $\pm$ 18,34 & 81,20 $\pm$ 31,55 & 139,60 $\pm$ 56,15 & 327,40 $\pm$ 31,85 & 398,20 $\pm$ 61,72 & 493,80 $\pm$ 91,81 & 398,40  \\
& \texttt{NF+S(2)} &48,00 $\pm$ 0,00 & 84,80 $\pm$ 9,39 & 210,20 $\pm$ 109,69 & 485,80 $\pm$ 101,44 & 581,60 $\pm$ 102,09 & 673,60 $\pm$ 90,09 & 581,80  \\
& \texttt{NF+S(4} &64,40 $\pm$ 22,46 & 122,60 $\pm$ 18,78 & 197,60 $\pm$ 27,12 & 511,40 $\pm$ 93,65 & 611,60 $\pm$ 83,38 & 753,60 $\pm$ 121,08 & 636,60  \\
& \texttt{NF+S(-2)} &\cellcolor{CornflowerBlue} \textbf{47,00 $\pm$ 0,00} & \cellcolor{CornflowerBlue} \textbf{59,40 $\pm$ 18,39} & \cellcolor{CornflowerBlue} \textbf{71,80 $\pm$ 9,07} & \cellcolor{CornflowerBlue} \textbf{270,40 $\pm$ 34,61} & \cellcolor{CornflowerBlue} \textbf{328,00 $\pm$ 37,58} & \cellcolor{CornflowerBlue} \textbf{456,00 $\pm$ 76,40} & \cellcolor{CornflowerBlue} \textbf{286,40}  \\ \cline{2-9}
& \texttt{IG} &84,20 $\pm$ 30,71 & 133,60 $\pm$ 73,34 & 154,20 $\pm$ 75,77 & 439,20 $\pm$ 177,27 & 513,60 $\pm$ 175,81 & 629,20 $\pm$ 175,41 & -  \\
& \texttt{IG+S(1)} &84,20 $\pm$ 72,02 & 117,40 $\pm$ 92,04 & 146,20 $\pm$ 80,36 & 345,00 $\pm$ 53,50 & 494,00 $\pm$ 66,06 & 597,00 $\pm$ 69,31 & 431,80  \\
& \texttt{IG+S(2)} &\textbf{80,00 $\pm$ 30,99}& \textbf{88,00 $\pm$ 28,99} & \textbf{108,80 $\pm$ 29,35} & \textbf{278,00 $\pm$ 61,11} & \textbf{381,60 $\pm$ 61,06} & 481,00 $\pm$ 70,55 & \textbf{340,20}  \\
& \texttt{IG+S(4)} &89,20 $\pm$ 162,95 & 130,60 $\pm$ 94,70 & 185,20 $\pm$ 63,57 & 327,00 $\pm$ 45,65 & 402,40 $\pm$ 38,89 & \textbf{477,40 $\pm$ 43,51} & 373,00  \\ \cline{2-9}
& \texttt{IG$^*$} &55,80 $\pm$ 18,57 & \textbf{109,60 $\pm$ 48,82} & \textbf{184,40 $\pm$ 47,66} & \textbf{388,20 $\pm$ 83,15} & \textbf{479,40 $\pm$ 142,33} & 720,80 $\pm$ 185,57 & -  \\
& \texttt{IG$^*$+S(2)} &\textbf{52,20 $\pm$ 9,39} & 118,40 $\pm$ 47,80 & 197,60 $\pm$ 53,84 & 435,40 $\pm$ 61,93 & 535,40 $\pm$ 91,63 & \textbf{636,00 $\pm$ 131,33} & \textbf{502,40}  \\
& \texttt{IG$^*$+S(4)} &288,56 $\pm$ 9,07 & 126,60 $\pm$ 31,12 & 222,40 $\pm$ 68,56 & 494,40 $\pm$ 97,42 & 590,00 $\pm$ 101,86 & 669,40 $\pm$ 133,25 & 510,60  \\ 
\hline
\multirow{9}{*}{$e_3$} & \texttt{NF} &\textbf{198,67 $\pm$ 43,99} & 484,17 $\pm$ 120,75 & \textbf{639,67 $\pm$ 171,15} & \cellcolor{CornflowerBlue}\textbf{1074,17 $\pm$ 253,13} & 1300,17 $\pm$ 264,92 & \cellcolor{CornflowerBlue}\textbf{1853,17 $\pm$ 311,92} & -  \\
& \texttt{NF+S(1)} &205,60 $\pm$ 41,26 & 529,40 $\pm$ 92,33 & 857,20 $\pm$ 108,38 & 1220,20 $\pm$ 197,36 & 1522,60 $\pm$ 241,32 & 1901,80 $\pm$ 436,33 & 1345,40  \\
& \texttt{NF+S(2)} &205,00 $\pm$ 41,33 & 567,60 $\pm$ 78,15 & 905,80 $\pm$ 281,01 & 1355,80 $\pm$ 246,25 & 1736,40 $\pm$ 316,32 & 2478,00 $\pm$ 504,59 & 1520,00  \\
& \texttt{NF+S(4)} &266,20 $\pm$ 76,58 & 680,80 $\pm$ 151,54 & 871,80 $\pm$ 208,36 & 1387,40 $\pm$ 301,58 & 1655,20 $\pm$ 259,20 & 2380,00 $\pm$ 678,52 & 1685,40  \\
& \texttt{NF+S(-2)} &246,80 $\pm$ 76,87 & \textbf{415,40 $\pm$ 94,84} & 749,00 $\pm$ 107,90 & 1091,60 $\pm$ 202,74 & \cellcolor{CornflowerBlue}\textbf{1273,00 $\pm$ 55,62} & 2592,60 $\pm$ 1177,43 & \cellcolor{CornflowerBlue}\textbf{1214,00}  \\ \cline{2-9}  & \texttt{IG} &204,40 $\pm$ 56,74 & \cellcolor{CornflowerBlue}\textbf{386,00 $\pm$ 109,12} & \cellcolor{CornflowerBlue}\textbf{584,00 $\pm$ 139,95} & 1758,20 $\pm$ 1057,47 & 1876,80 $\pm$ 1026,90 & 2646,00 $\pm$ 1088,98 & -  \\
& \texttt{IG+S(1)} &\cellcolor{CornflowerBlue}\textbf{179,20 $\pm$ 34,51} & 403,80 $\pm$ 73,61 & 589,00 $\pm$ 54,46 & \textbf{1090,80 $\pm$ 180,15} & \textbf{1285,60 $\pm$ 229,85} & \textbf{1890,00 $\pm$ 563,90} & \textbf{1216,40}  \\
& \texttt{IG+S(2)} &247,00 $\pm$ 77,02 & 392,60 $\pm$ 71,53 & 689,00 $\pm$ 157,87 & 1174,20 $\pm$ 145,49 & 1342,20 $\pm$ 155,32 & 1886,20 $\pm$ 307,13 & 1385,20  \\
& \texttt{IG+S(4)} &204,80 $\pm$ 82,56 & 386,00 $\pm$ 117,33 & 688,20 $\pm$ 93,78 & 1029,60 $\pm$ 221,93 & 1327,60 $\pm$ 302,20 & 2521,80 $\pm$ 356,76 & 1397,20  \\ \cline{2-9}
& \texttt{IG$^*$} &285,40 $\pm$ 67,76 & 547,60 $\pm$ 118,51 & \textbf{792,80 $\pm$ 131,78} & 1476,60 $\pm$ 405,61 & 1661,20 $\pm$ 523,74 & 2785,80 $\pm$ 840,53 & -  \\
& \texttt{IG$^*$+S(2)} &\textbf{269,20 $\pm$ 100,18} & 626,60 $\pm$ 166,32 & 897,60 $\pm$ 204,44 & 1433,00 $\pm$ 256,23 & 1682,00 $\pm$ 250,60 & 2666,80 $\pm$ 429,18 & 1711,80  \\
& \texttt{IG$^*$+S(4)} &274,00 $\pm$ 76,59 & \textbf{519,60 $\pm$ 49,73} & 848,40 $\pm$ 152,19 & \textbf{1353,20 $\pm$ 273,89} & \textbf{1579,20 $\pm$ 292,15} & \textbf{2286,20 $\pm$ 410,90} & \textbf{1428,00}  \\\hline
\end{tabular}}
\caption{Area covered $A$ against time for different strategies. In \textbf{bold} the best value per exploration strategy, and in \textcolor{CornflowerBlue}{blue} the best value overall for each environment.}
\label{tab:results}
\end{table*}
%\red{Results:
%\begin{itemize}
%    \item description of setup
%    \item the three maps
%    \item ROS config
%    \item PC specs
%    \item comparison wrt different values of beta
%    \item negative betas
%\end{itemize}
%}

\begin{figure}[t]
\centering
\subfloat[$e_2$\label{fig:map2}]{\includegraphics[trim={0cm 0cm 0 0cm},clip,width=0.45\linewidth]{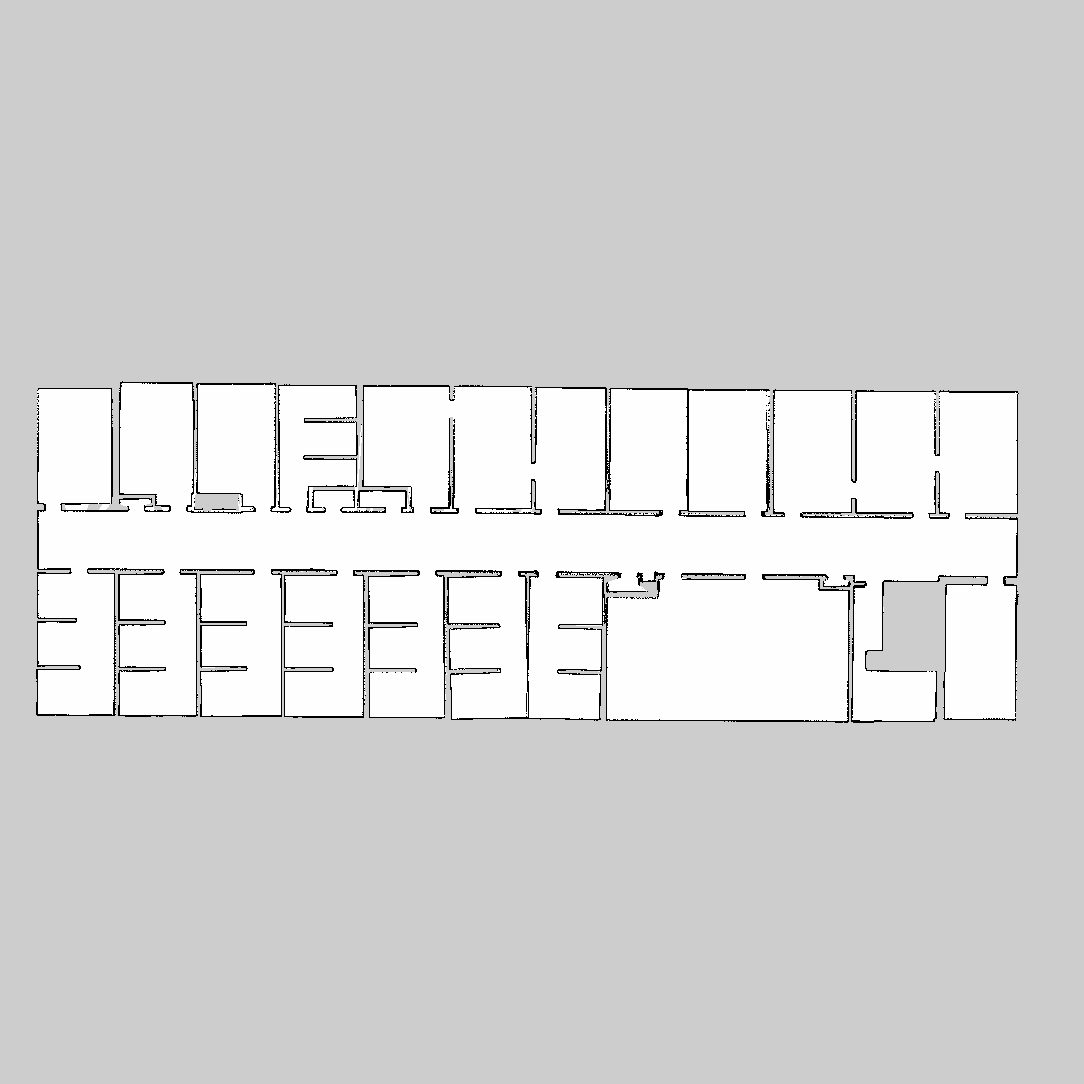}}
\subfloat[$e_3$\label{fig:map3}]{\includegraphics[trim={0 0cm 0 0cm},clip,width=0.45\linewidth]{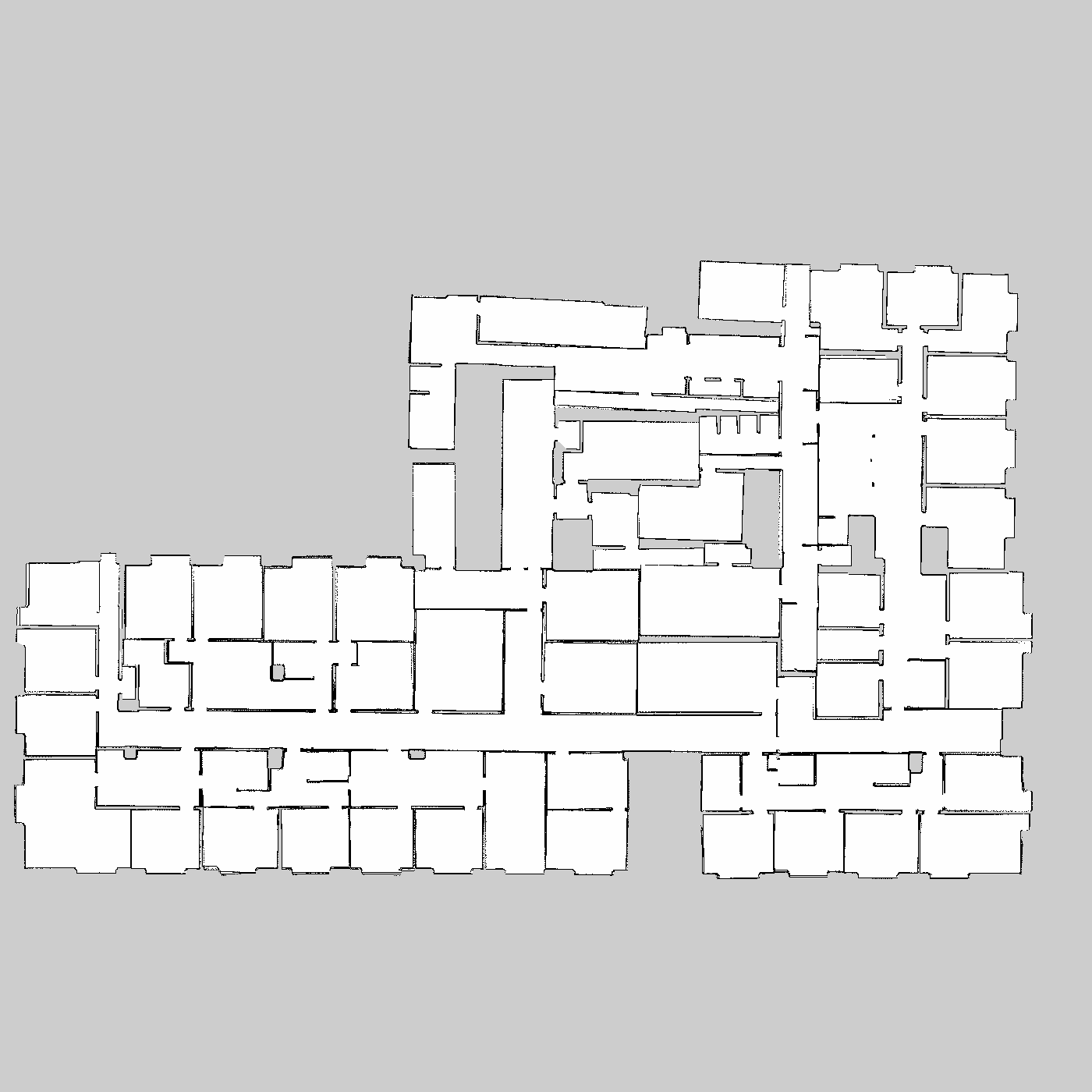}}
\caption{Two of the three environments used for experiments.}
\label{fig:exp/moremaps}
\end{figure}

%\red{The robot wants to follow an optimal solution, which is unknown. The exploration strategy is trying to replicate the behaviour of this optimal strategy, but without the same information. In some environments, the optimal solution is similar to methods that follow an IG-based solution, so quick exploration first; in other environments, the optimal solution is to greedily explore locally the environment and then move on. Our method, when tuned, can be used to bias the strategy towards the unknown optimum, but this is somehow dependent on the environment.}

In this section, we perform an extensive experimental evaluation of how using saliency areas can inform different exploration strategies.

%\red{More precisely we use different values of beta, what is the caveat in this?}

\begin{figure}[t]
\centering
\includegraphics[width=0.75\linewidth]{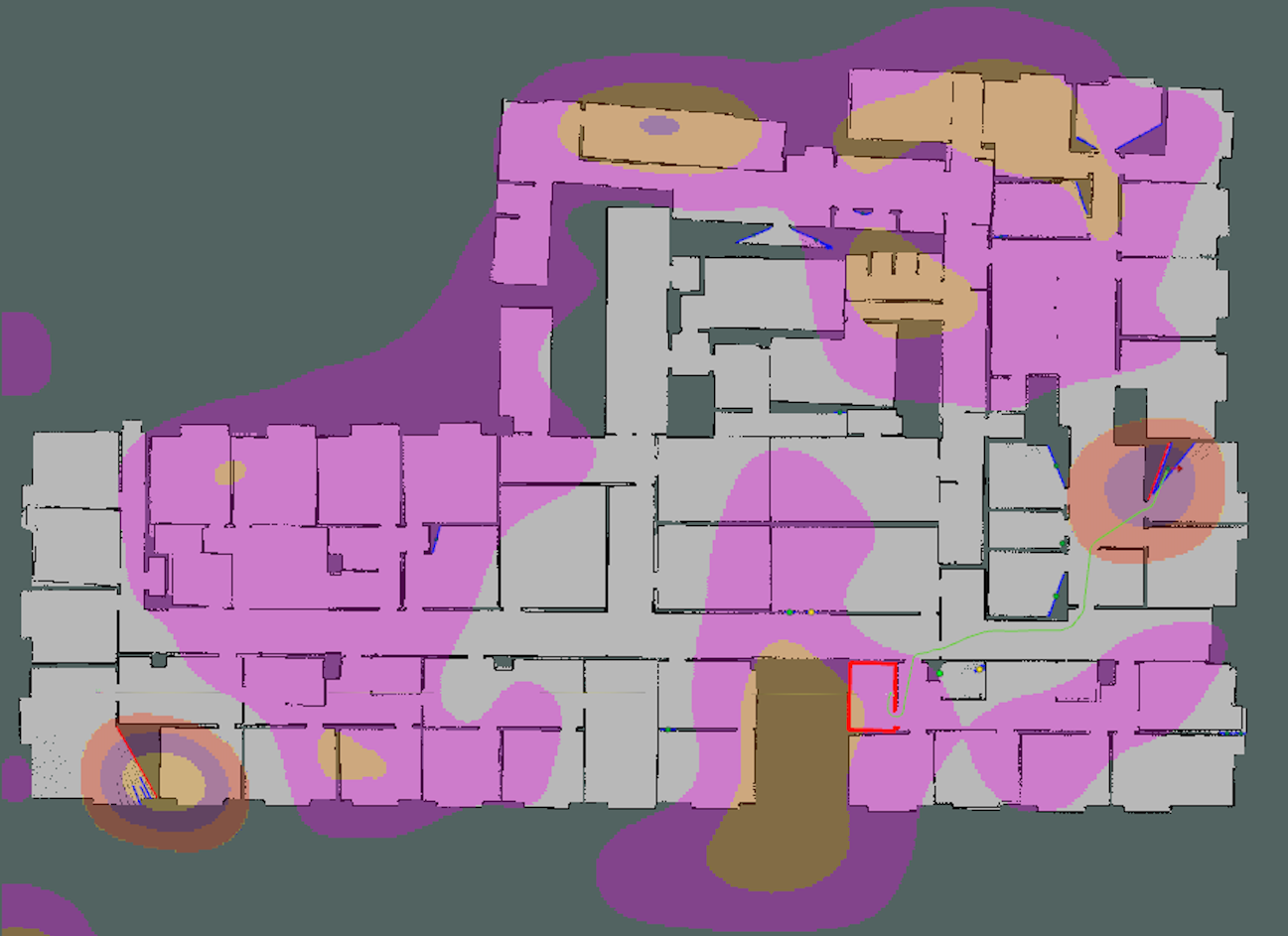}
\caption{An example of our system running in ROS of a map considered as \texttt{explored} by our termination criterion. In this case, the saliency map from the last \texttt{not-explored} map is used to calculate $S()$.}. 
\label{fig:rviz}
\end{figure}

\subsection{Experimental setting}\label{sec:expsettings}

We use three different simulated large-scale environments from the dataset representing the MIT+KTH campuses \cite{amigoni2018improving}, whose maps are shown in Fig. \ref{fig:map1} (environment $e_1$) and Fig. \ref{fig:exp/moremaps} (environments $e_2$ and $e_3$.) %\textcolor{red}{WE MUST SAY WHICH ENEVIRONMENT IS CALLED $e_1$, $e_2$, AND $e_3$, RESPECTIVELY. REMEMBER TO CHECK REFERENCES TO FIGURES.}
We use ROS, with a Turtlebot3 robot simulated in STAGE, and equipped with a 2D LiDAR with \SI{10}{\meter} of range and a \ang{360} field-of-view, with realistic simulated noise on sensor readings and robot motion. We use~\cite{gmapping2007tro} for SLAM and the ROS navigation stack for navigation. Experiments are performed on an i7-7700 CPU and a 1060 GPU, on a computer with an i7-7400 CPU and a 1050 GPU, and on a Ryzen7 5700 CPU with a 3060Ti GPU.  
The Grad-CAM from \cite{luperto2024estimating} processes the map in real-time and is integrated with the ROS navigation stack.

We measure the progression (over time) of the percentage of covered area $A$ and distance travelled by the robot. For the sake of brevity, we report only the time as it is proportional to distance. We indicate with $A_x$ the time required to cover $x\%$ of the total area of the map. We also record the time $\bar{T}$ when the termination criterion from \cite{luperto2024estimating} is triggered. We do 5 runs in each environment for each configuration, reporting average and standard deviation.
We continue the exploration until all reachable frontiers are explored.
Saliency maps are available only when the termination criterion from \cite{luperto2024estimating} has not been triggered, as they are obtained for maps labelled as \texttt{not-explored}. After a map is labelled as \texttt{explored}, we use the last available saliency areas, as they still indicate areas where promising frontiers are located. An example is shown in Fig. \ref{fig:rviz}.

\subsection{Experiemental results}

\begin{figure}[t]
\centering
\includegraphics[width=\linewidth]{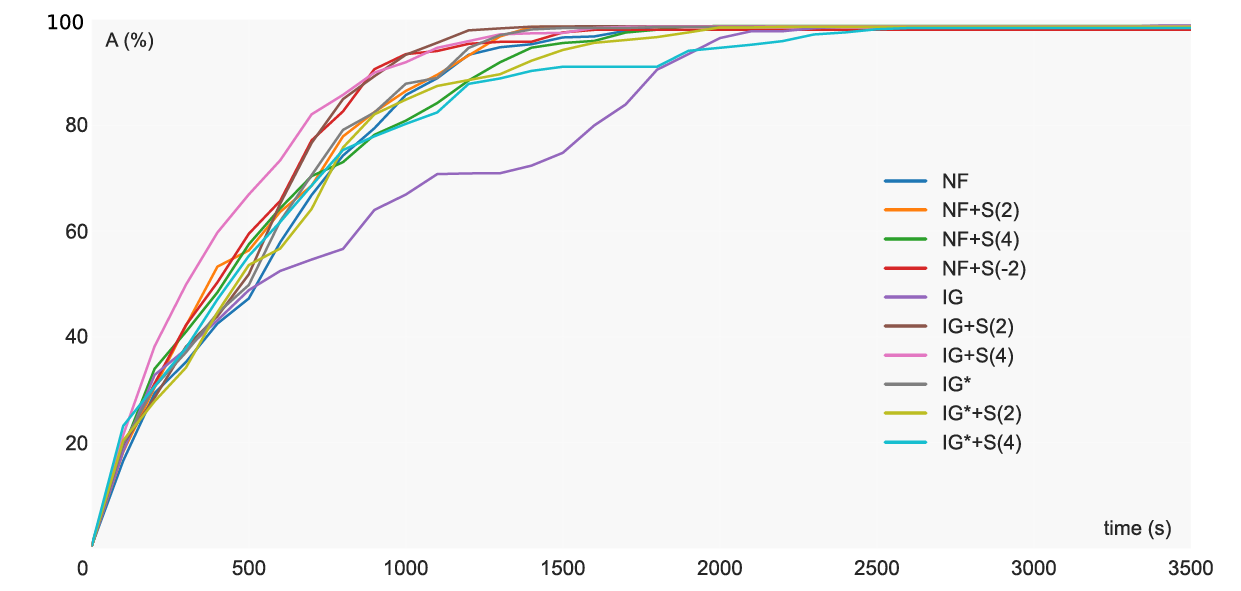}
\caption{Progression of explored area against time for $e_1$.}
\label{fig:progression}
\end{figure}

%\textcolor{red}{IS FIG. 7 MENTIONED IN THE TEXT? CHECK.}
We evaluate how different values of $\beta$ in Eq. \ref{eq:2} influence the performance of the three different exploration strategies of Section \ref{sec:strategies}. We indicate with \texttt{+S($\beta$)} an exploration strategy that uses saliency maps with a given value for $\beta$. Table \ref{tab:results} shows the results; we highlight in \textbf{bold} the best value per exploration strategy, and in \textcolor{CornflowerBlue}{blue} the best value overall. Fig. \ref{fig:progression} shows the progression over time of $A$ for $e_1$. %\textcolor{red}{IS THIS CLAIM CORRECT? I DON'T SEE DETAILS IN FIG. 2.}

%In Eq. \ref{eq:1}, the two components follow two competing objectives. The aim of $dist(f)$ is to push the robot to perform an exploration step-by-step, thus exploring its local neighbourhood thoroughly before going to new areas (e.g., new rooms). The second component, $IG(f)$, pushes the robot to new areas of the environment, with the desiderata of having a high increase in the mapping area, at the expense of leaving unexplored areas behind. The combination of the two factors balances between these two competing objectives.
%In the following sections, we describe how different parameters of $\beta$ bias these two drives.

We test different values for $\beta = \{-2,1,2,4\}$. As saliency areas represent portions of the environment that are important to explore, a positive value of $\beta$ in Eq. \ref{eq:2} is biasing the robot to maximise the explored area in the short term, thus pushing the robot towards unexplored portions of the environment.  Consequently, positive values of $\beta$ reinforce the aggressive behaviour triggered by the $IG$. Conversely, negative values of $\beta$ push the robot to fully explore its neighbourhood first, setting up a cost for the robot to explore frontiers with high saliency, namely frontiers that can lead to explore large unvisited parts of the environment.

Overall, results from Table \ref{tab:results} show that the use of saliency areas can have a relevant, positive effect on all exploration strategies, as it can allow the robot to detect which frontiers lead to highly-promising unexplored areas of the environment from those that are not.

When we consider \texttt{NF}, positive values of $\beta$ allows the robot to have a behaviour similar to those of methods using $IG$, thus allowing the robot to reach faster higher percentage of explored areas (up to $A_{70}$), but paying the price of having to backtrack in later stage of exploration.

More interestingly, a negative value of $\beta=-2$ often allows the robot to fully explore the entirety of the environment in less time, and it is often the best exploration strategy among the options we tested. 
This is because \texttt{NF} always selects the closest frontiers, but this can lead the robot to move far from its current location (e.g., if the closest frontier is a doorway leading the robot to a new room). With \texttt{NF+S($-2$)}, we penalise those frontiers that lead the robot far away if they open the way to new parts of the environment, thus promoting a more fine-grained exploration of its immediate vicinity, and thus furtherly increasing the “conservative" behaviour of \texttt{NF}. These findings confirm those of \cite{ericson2025information}, which recently presented an exploration strategy that explore first the local vicinity of the robot with the effect of minimizing the future distance covered by the robot to explore other frontiers. This can be seen in particular in the environments $e_1$ and $e_2$. The other environment ($e_3$) is particularly complex, and thus the behaviour of \texttt{NF} and \texttt{NF+S($-2$)} is more similar.

When we consider exploration strategies that use $IG$, we test only positive values of $\beta$, as $IG$ and $S(f)$ are correlated and both push the robot towards the same desiderata of selecting highly relevant frontiers first. Negative values of $\beta$ would compete with $IG$ and override it.  We can see how \texttt{IG+S($2$)} and \texttt{IG+S($4$)} improve in early stages of exploration runs and allow the robot to reach up to $A_{90}$ in less time than \texttt{IG}. As expected, this behaviour has the cost that the robot is forced to backtrack in later stages of exploration, thus requiring higher times to reach $A_{99}$ when compared with \texttt{NF}.

Finally, for the exploration strategy that knows the real information gain of each frontier, \texttt{IG$^*$}, adding bias to highly relevant frontiers does not improve results. This is because the saliency map $S(f)$ is an estimation of promising areas, while $IG^*(f)$ already provides the true information gain that the robot obtains by visiting the frontier. However, it is also interesting to observe how using an estimated $IG$ with a high value of $\beta$ (e.g.,  \texttt{IG+S($4$)}) is often a better choice even if the robot has the full knowledge of the real $IG$, as in \texttt{IG$^*$}. 
This furtherly proves how saliency maps can be seen as a more fine-grained method to evaluate the information gain of frontiers, as they estimate the fact that frontiers can lead in the medium-long term to observe entirely new portions of the environment. Finally, we note that the use of the termination criterion allows for a significant reduction in the total exploration time $\bar{T}$, especially when \texttt{NF+S(-2)} is used.

%In general noi non sappiamo quale sia strategic migliore, che dipende sia dai desiderata che dal tipo di ambient. ma potrebbe anche essere che per alcuni ambient una strategia molto locale sia migliore di una molto greedy.

%\subsubsection{Biasing Nearest Frontier Strategies}
%Effetti su strategies che prediligono una exploration locale: saliency cerca di portarli lontano e quindi da un certo punto di vista porta dei vantaggi nel limitare il loro behaviour molto greedy. però una cosa interessante è che dare un bonus negative, ovvero sfavorire quelle frontier piu promettenti the portano il robot altrove riesce a rendered il comportamento del robot ancora piu conservative e quindi alla fine ad avere risultati migliori; il robot in questto modo prima finisce di esplorare il locale poi va via. Difatto il "buono" delle strategies greedy è che non considerano IG; le saliency sono un concetto + raffinato di IG e quindi aggiungerle di fatto "sposta" queste strategies verso quelle IG-based. solo che IG sposta il tipo di behaviour.

%\subsubsection{Biasing Information-Gain Strategies}
%Effetti su strategies che prediiligono una esplorazione veloce: quando IG è stimato il boost è evident perchè permette di scegliere nelle frontier a high ig quelle che sono migliori e quindi andare direttamente li. Invece per quanto riguarda il GT perfetto, siccome saliency sono approssimate quello che fa il metodo difatto è peggiorare un tipo di knowledge che gia il robot ha e giustamente quindi non migliorano le performance o talvolta le peggiorano.

%NOTA su stopping criterion
\section{Conclusion}
In this work, we presented a method to estimate the most promising areas to visit in a partially explored environment, to move the robot towards a faster completion of the exploration. We have shown how these saliency areas can inform different exploration strategies to meet different desiderata.%: adding a bonus to high-values of saliency areas can led to a faster exploration in the short term, while adding a cost to those area encourages the robot to explore first its local neighborhood, thus allowing the coverage of the full map in shorter time. 

Future works involve testing the concept of saliency map in combination with map predictive methods\cite{ho2024mapex} and with learning-based approaches~\cite{cao2023}, as well as an evaluation with real robots.

\bibliographystyle{IEEEtran}
\bibliography{citations}

\end{document}